\documentclass{article} 
\usepackage{arxiv}

\usepackage[american]{babel}
\usepackage{amssymb}
\usepackage[titletoc]{appendix}
\usepackage{dblfloatfix}
\usepackage{wrapfig}

\usepackage{natbib} 
\usepackage{mathtools} 
\usepackage{booktabs} 
\usepackage{tikz} 

\usepackage[linesnumbered, ruled, vlined]{algorithm2e}
\usepackage{algpseudocode}

\title{Non-stationary Dynamic Pricing via Actor-Critic Information-Directed Pricing}

\author{
    Po-Yi Liu\\
	Department of Statistics\\
	National Chengchi University\\
	Taipei, Taiwan \\
	106305022@g.nccu.edu.tw \\
	\and
	\textbf{Chi-Hua Wang}\\
	Department of Statistics\\
	University of California\\
	Los Angeles, CA, USA \\
	chihuawang@ucla.edu \\
    \and
    \textbf{Henghsiu Tsai}\\
    Institute of Statistical Science\\
    Academia Sinica\\
    Taipei, Taiwan\\
    htsai@stat.sinica.edu.tw \\
    }

\begin{document}
\maketitle
\begin{abstract}
This paper presents a novel non-stationary dynamic pricing algorithm design, where pricing agents face incomplete demand information and market environment shifts. The agents run price experiments to learn about each product's demand curve and the profit-maximizing price, while being aware of market environment shifts to avoid high opportunity costs from offering sub-optimal prices. The proposed \texttt{ACIDP} extends information-directed sampling (IDS) algorithms from statistical machine learning to include microeconomic choice theory, with a novel pricing strategy auditing procedure to escape sub-optimal pricing after market environment shift. The proposed \texttt{ACIDP} outperforms competing bandit algorithms including Upper Confidence Bound (UCB) and Thompson sampling (TS) in a series of market environment shifts. 

\end{abstract}

\vspace{-2mm}
\section{Introduction}
\vspace{-2mm}

We consider a non-stationary dynamic pricing problem, a setting in which pricing agents run pricing experiments to learn incomplete demand information to cover profit-maximizing price (dynamic pricing) while being mindful of high opportunity costs from offering sub-optimal price due to market environment shift (non-stationary). Such a setting arises naturally in industrial practice but remains unanswered in the literature, preventing practitioners from deploying state-of-the-art online decision-making methodology effectively in modern online service industries.

Learning incomplete demand information from pricing experiments is in emerging demand from industrial practice due to the opportunity cost accompanied with sub-optimal price and inefficient exploration.\citep{misra2019dynamic}. Unfortunately, even when algorithms find optimal prices successfully, the offered price degrades to sub-optimal once the market environment shifts. Therefore, it is desirable to systematically detect the environmental changes, abolish the running pricing strategy that appears to be unpromising, and turn to collect the latest market information. Regrettably, though optimality had been well studied in the existing designs of multi-armed bandit (MAB) algorithms including Upper Confidence Bounds (UCB) \citep{lai1985asymptotically, auer2002finite, garivier2011kl, maillard2011finite, kaufmann2012bayesian}, Thompson Sampling (TS) \citep{thompson1933likelihood, granmo2010solving, scott2010modern, chapelle2011empirical, may2011simulation, russo2018tutorial, russo2016information} and Information Directed Sampling (IDS) \citep{russo2018learning, kirschner2020information, kirschner2021asymptotically}, they failed to \textit{transfer} their learned decision mechanism into the shifted market environment. In modern dynamic pricing practice, deployment of MAB methodology turns out to be impeded by such transfer inability. Resolving \textit {transfer inability} is crucial to the advancement of practical dynamic pricing policy.

Online hypothesis testing is the key statistical instrument to assess and monitor the credibility of learned demand information. These test statistics measure the discrepancy between the learned and current market environment model, suggesting possible pricing optimality disruption. The fundamental rationality of such alarm is to utilize the unlikely occurrence of large deviations given the null hypothesis, the non-existence of environmental changes. Yet, these tools are far from both dynamic pricing and bandit literature, encouraging our investigation of the non-stationary dynamic pricing problems.

Solving this fundamental problem based on existing knowledge of MAB pricing \citep{misra2019dynamic} requires overcoming three major challenges:

\textbf{(a) Market environment shift.} Depart from optimality in the stationary environment of existing bandit algorithm design, our approach should allow environment shift alarm and quick reaction to change sub-optimal price to the new optimal price 
to reduce regrettable opportunity cost.

\textbf{(b) Indirect customer feedback.} Depart from independent assumption between available arms in multi-armed bandit problems, our approach should utilize microeconomic choice theory to infer indirect customer feedback through strategical pricing experiments and incorporate such indirect information into the pricing pipeline.

\textbf{(c) Pricing performance monitoring.} Being aware of potential market environment shifts, our approach should build a statistical method-based auditing pipeline to monitor the pricing performance to ensure the transferability of the pricing decision mechanism in the current market environment.

\vspace{-3mm}
\subsection{Main contributions}
\vspace{-3mm}

\begin{figure*}
  \centering
  \includegraphics[width=1\textwidth]{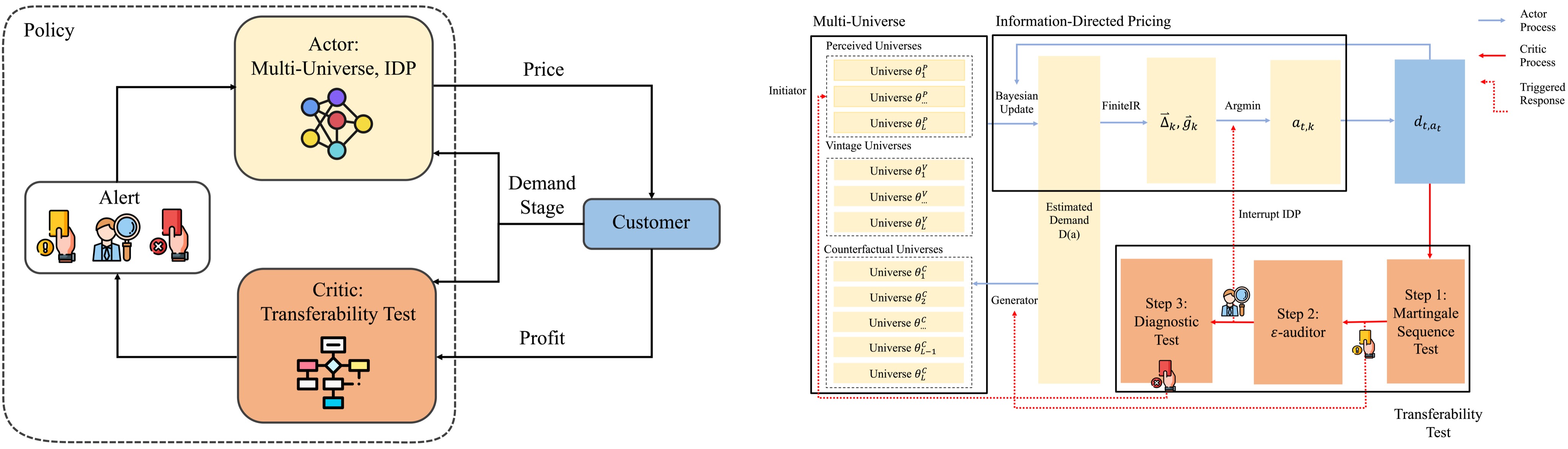}
  \caption{\texttt{ACIDP}:\textbf{Actor-Critic Information-Directed Pricing}.}\label{fig:pipe_algo}
\end{figure*}

We summarize the knowledge advancement of this work contributing to existing bandit-based dynamic pricing algorithm design (that learns a stationary-environment-optimal pricing policy without awareness of potential market environment change): 

\textbf{(I) \texttt{ACIDP} non-stationary dynamic pricing algorithm. (Figure \ref{fig:pipe_algo}, left)} 
We deliver the Actor-Critic Information Directed Pricing (\texttt{ACIDP}) algorithm that learns profit-maximizing price via incomplete demand information and is able to transfer pricing optimality in the presence of market environment shift.
The key of pricing mechanism transfer ability inside \texttt{ACIDP} is to construct the \textit{Multi-Universe}, a space of synthetic bandit environments shaped by environment exploration, exploitation, and experimentation. By evaluating the expected profit and demand information in each universe of Multi-Universe, we adopt the Bayesian rule to maintain a posterior belief distribution over universes to implement the well-known information directed sampling (IDS) \citep{russo2018learning} for pricing decision. Then, we develop an auditing pipeline to ensure pricing optimality transferability.

\textbf{(II) Transferability auditing pipeline (Figure \ref{fig:pipe_algo}, right)} We design a transferability auditing pipeline to monitor market environment changes and inform pricing agents if the credibility of the learned consumer demand model is no longer accurate enough to inform profit-maximizing price. The auditing pipeline consists a 3 stage sequence of events: yellow card (credibility inspection; Section \ref{subsec:pricing_strategy_auditing}.A), audit sampling (optimality inspection; Section \ref{subsec:pricing_strategy_auditing}.B) and red card (transfer-ability inspection; Section \ref{subsec:pricing_strategy_auditing}.C). This auditing pipeline contributes a novel framework in dynamic pricing literature.

\textbf{(III) Empirical results.} Via a combination of synthetic market environments (stationary, structure shift, upside-down and real-world scenario), we show the success of \texttt{ACIDP} is rooted in (a) benefits of synthetic bandit environments and (b) benefits of auditing pipeline. The empirical results support the superiority of \texttt{ACIDP} over the existing bandit-based pricing method. We show the established environment synthesis pipeline and environment auditing pipeline leads to substantial regret reduction across different market environment shifts.

\vspace{-3mm}
\subsection{Related works}
\vspace{-3mm}

\textbf{Bandit-based dynamic pricing and demand learning.} Dynamic pricing is acknowledged to be challenging without knowing demand function \citep{besbes2009dynamic}. While MAB offers a cutting-edge partial monitoring setup\citep{cohen2021correlated, chen2021multi}, bandits literature on dealing with correlated and non-stationary rewards are still far from practical use.\citep{gupta2018exploiting, besbes2014stochastic,  slivkins2008adapting, gupta2020correlated}. Our framework is based on \citep{misra2019dynamic}, where an upper confidence bound-based dynamic price experimentation policy was proposed under the stationary assumption and extends MAB to microeconomic choice theory. Our work further investigates the performance of Thompson sampling and Information Directed Sampling and identified the transfer inability in these methods when the market environment shifts. Our work adds to the dynamic pricing literature by developing a non-stationary pricing method.
\\ \hspace*{\fill} \\
\textbf{Bandit algorithms and Information directed sampling.} Information directed sampling (IDS) \citep{russo2018learning} is the state-of-the-art bandit policy design principle. While IDS finds several applications in bandit learning literature \cite{kirschner2020information, kirschner2021asymptotically}, it remains open in the literature of dynamic pricing. Intuitively, IDS could utilize information structures of dynamic problems such that sampling one price can inform the pricing agent's assessment of other prices. Our work contributes a new bandit problem with non-stationary rewards and a novel implementation of IDS in dynamic pricing with optimality auditing.
\\ \hspace*{\fill} \\
\textbf{Online hypothesis testing and confidence sequence.} Our auditing procedure borrows the modern and classical statistical tools: martingale confidence sequence and hypothesis testing. The yellow card test in our auditing pipeline \citep{maillard2019mathematics, howard2020time, howard2021time} gives generic construction of martingale confidence sequence, which is a suitable statistical tool to alarm environment shift. For the red card test in our auditing pipeline, we adopt an exact test that quantifies the statistical significance of deviations from a theoretically expected distribution of observations using sample data. Such property conforms to the need of identifying the fitness of the demand curve. Our work contributes to statistics literature with an innovative application as an auditing pipeline.
\vspace{-3mm}
\subsection{Problem formulation and notation}
\vspace{-3mm}
This section defines dynamic pricing under partially identified demand learning in \cite{misra2019dynamic}.
\\ \hspace*{\fill} \\
\textbf{(I) Price, its demand, and its profit.} At pricing step, the agent choose the price $a$ from a finite set of $K$ prices $\left\{a_{1}, \ldots, a_{K}\right\}$. Each price $a$ associates with its true demand $D(a)$, which is unknown and to be learned during the dynamic pricing process. The true profit of price $a$ is defined as $R(a)=aD(a)$. 
\\ \hspace*{\fill} \\
\textbf{(II) Batch pricing experiment, pricing policy and observed demand/profit.} We assume prices can change after every $N$ consumers who visit the product. That is, each pricing step consists of a batch of $N$ consumers. A dynamic pricing algorithm, $\Psi$, selects prices based on the history of past prices and earned profits; formally, given the filtration $H_{\tau}=\left\{a_{\tau}, d_{\tau, a_\tau} \mid \tau=1, \ldots, t-1\right\}$, $a_{t}=\Psi\left(H_\tau\right)$. The observed demand $d_{t}(a_{t})$ is the observed number of purchase after offering price $a_{t}$ at pricing step $t$; formally, the observed demand is defined as $d_{t}(a_{t})=\sum_{i=1}^{N}I(v_{i,t}\geq a_{t})$, where $v_{i,t}$ is the valuation the $i$th customer in current pricing batch willing to pay and the consumer make purchase if and only if the offered price $a_{t}$ is not greater than $v_{i,t}$. The observed profit of price $a_{t}$ is $r_{t}(a_{t})\equiv a_{t}d_{t}(a_{t})$.
\\ \hspace*{\fill} \\
\textbf{(III) Objective of dynamic pricing policy.} Our goal is to design pricing policy $\Phi$ such that the offering price sequence $a_{1}, a_{2}, \cdots, a_{T}$ maximize the cumulative observed profit  $\sum_{t=1}^{T}r_{t}(a_t)=\sum_{t=1}^{T}a_{t}d_{t}(a_{t}).$ And in bandit terminology, the performance is measured by minimizing the regret $\Delta_T(\Phi) = \mathbb{E}[\sum^T_{t=1}r_t(a^*_t)] - \sum_{t=1}^{T}r_{t}(a_t)$ where $a^*_t=\arg\max_{a\in \{a_1,\cdots, a_{K}\}}R(a)$ is the optimal price at $t$ that maximizes the profit.
\vspace{-3mm}
\section{Dynamic Pricing in Non-stationary Environment}
\vspace{-3mm}
\label{sec:non-stat_DP}

We assume at time $t$ underlies a customer valuation distribution such that $v_{i,t} \overset{i.i.d.}{\sim} f_{V_t}$. The market environment shifts when $f_{V_{t+1}} \neq f_{V_t}$. The shifting moment and transformation is assume to be unknown. This section elaborates dynamic pricing problems in the presence of such environment shifts from three perspective: desideratas, challenges and our resolutions.
Section \ref{subsec:desirata} depicts the three desideratas of dynamic pricing algorithms: (a) credibility, (b) optimality and (c) tranfer-ability. 
Section \ref{subsec:challenge} identifies the three major challenges from market environment shifts: (a) credibility degradation, (b) optimality disruption and (c) transfer inability. 
Section \ref{subsec:resolution} delivers our framework that resolved these challenges: (a) universe synthesizer (b) information-directed pricing and (c) pricing strategy auditing.
\vspace{-3mm}
\subsection{Desiderata of pricing algorithms}
\vspace{-3mm}

\label{subsec:desirata}

Our goal is to design a pricing algorithm $\Psi$ that monitors the market environment in order to \textbf{(I) evaluate} the optimality of earned profit of the current pricing policy $\Psi$, and \textbf{(II) audit} the pricing policy $\Psi$ ``transferability`` by testing whether the current learned consumer demand curve sufficient to offer optimal pricing. In order for the pricing algorithms $\Psi$ to fulfill the evaluation and auditing tasks, it must satisfy the following desiderata: \textbf{(I)} it should be able to \textbf{identify optimal price distribution shift} through \textit{statistical} measures of earned profit, and \textbf{(II)} it should be \textbf{decision-wise computable,} i.e., we should be able to tell if any given individual pricing decision is no longer optimal due to market environment shift. 
\\ \hspace*{\fill} \\ 
Having outlined the desiderata for dynamic pricing policy, we now propose three qualities that the policy $\Psi$ should be equipped with. $\Psi$ falls behind if it fails to fulfill all of the three qualities corresponding to the challenges depicted. These qualities are: 

\vspace{-2mm}
\begin{enumerate}
\item \textit{Credibility}--the learned demand model covers real consumer behavior from the market environment. A credible learned demand model should resemble ``realistic`` consumer transaction behavior.
\item \textit{Optimality}--the pricing decision are greedy enough to earn the best-possible profit, i.e. a pricing algorithm should be able to offer a near-optimal price as if the true consumer demand model is available. 
\item \textit{Transferability}--the pricing mechanism should not degenerate confronting market environment shifts, i.e., algorithms that converge to stationary environment optimal price is not truly "transferable".
\end{enumerate}
\vspace{-3mm}

\vspace{-3mm}
\subsection{Challenges from market environment shifts}
\vspace{-3mm}
\label{subsec:challenge}

Thinking in Bayesian language, the three modes of inabilities correspond to defection in posterior (A. model credibility degradation), likelihood (B. price optimality disruption), and prior (C. monitor transfer inability).

\textbf{A. Model credibility degradation. (Posterior)}
The market environment shifts drive changes in demand curves. Traditional stationary policies models uncertainty of the environment on separate arms with the underlying assumption that all arms share a single static environment. However, that means the uncertainty of each arms' result needs to be updated separate. Thus, the performance is injured while judging against the truth basing on biased information. In \ref{sec:exp_to_imple}, we show how policies behave differently going from a static environment to a dynamic environment. For instance, in periodic market shifts (Case4), UCB and TS only recognize optimal price at the beginning of the period as they are dragged by earlier faith.


\textbf{B. Pricing optimality disruption. (Likelihood)}
The shifted demand curve also changes the optimal price. Once the environment shifts, the likelihood of demand that the algorithm had been exploiting is obsolete, uninformative to inferring the shape of the profit curve. Such misinformation results in selecting sub-optimal prices, vandalizing the optimality of pricing policy.


\textbf{C. Monitoring Transfer inability. (Prior)}
Our current belief on the nature affects our decision. While Transfer-ability is a common issue in machine learning. failing to react to a new environment directly yields considerable loss in dynamic pricing problems. If the market environment shift is confirmed, then resetting belief is necessary for a valid non-stationary algorithm, enhancing the credibility of modeling and reinforcing the optimality of pricing. Contrarily, stationary policies jump into wrong conclusions in upwards trend and become inconclusive in downwards trend. See demonstration in Section 4.2.


\vspace{-3mm}
\subsection{Resolutions to non-stationary dynamic pricing}
\vspace{-3mm}
\label{subsec:resolution}

\textbf{A. Enhance credibility by universe synthesizer.}
Section \ref{subsec:env_synth} The universe synthesizer aids information-directed pricing by consistently providing bountiful bandit environments to explore from. We identify three data sources. 1. Direct samples that are costly but effective. 2. Past data that are useful but not necessarily accessible. 3. Synthetic data that expands the previous two and does not require additional exploration. The universe synthesizer is in charge of allocating the least rounds by manipulating the resources and offer IDP as much information as possible.
\\ \hspace*{\fill} \\
\textbf{B. Pursue optimality via information-directed pricing.}
Section \ref{subsec:InfoDir_pricer} implements information-directed sampling. ACIDP strikes a balance between expected regret and expected information gain of each price. In Bayesian formulation, we model the above-discussed universes as a random variable $\theta$ each, such that conditioned on $\theta$, $d_\tau$ is an iid sequence. $p(\theta)$ not only helps capture our belief that reflects the true nature, but allows ACIDP to quantify the expected regret and mutual information between $a^*$ and $d_\tau$.
\\ \hspace*{\fill} \\
\textbf{C. Monitor transferability via pricing strategy auditing.} Section \ref{subsec:pricing_strategy_auditing}
Price auditing assesses a firm's price policy. It helps ensure that prices offered are aligned with the consumer's willingness to pay. The audition consists of: a martingale sequence prescreening, an $\epsilon$-greedy based explorer, and a diagnostic test. The first one detects the anomaly and summons the universe synthesizer to expand resources while exploring a little with $\epsilon$-greedy. Lastly, the aforementioned exploration helps diagnose environmental changes via hypothesis testing.

\vspace{-3mm}
\section{Methodology}
\vspace{-3mm}

\textbf{Actor-Critic Information-Directed Pricing (\texttt{ACIDP})}
Algorithm 1 presents our solution, the \texttt{Actor-Critic Information-Directed Pricing} algorithm, to the non-stationary dynamic pricing problems established at Section \ref{sec:non-stat_DP}. There are 3 key components in our design of \texttt{ACIDP} algorithms: Universe Synthesizer, Information directed pricing, and Transferability Test. Section \ref{subsec:env_synth} establishes the Universe synthesizer to construct candidate universe to capture system uncertainty of the market environment, as solutions to the challenge \ref{subsec:challenge}.A. Section \ref{subsec:InfoDir_pricer} illustrates how \texttt{ACIDP} explores the synthetic environments supported by information directed sampling principle to maximize the cumulative profit, as a solution of challenge \ref{subsec:challenge}.B. Section \ref{subsec:pricing_strategy_auditing} elaborates the transferability test to audit environment stationarity that allow \texttt{ACIDP} a quick reaction when face with environment shift, as a solution of challenge \ref{subsec:challenge}.C.

\begin{algorithm}
\label{alg:ACIDP}
\caption{Actor-Critic Information-Directed Pricing}
\SetAlgoLined
\DontPrintSemicolon
\For{$t$ from 1 to $T$ do}{
$\vec{\Delta}$,  $\vec{g}$ $\leftarrow$ finiteIR$(L, K, N, p, q)$\;
         $a_t$ $\leftarrow$  $a_k$, $k \in \arg\min_k$ $\vec{\Delta}_k^2/\vec{g_k}$\;
    \If{alert = yellow card}{
        $q_{\theta^C}$ $\leftarrow$ Generator$(c)$\;
        $p(\theta^C_i)$ $\leftarrow$ $1$,
        $p(\theta)$ $\leftarrow$
        $\frac{p(\theta)}{\Sigma_{\theta}{p(\theta)}}$\;
        Sample $e \sim$ $bernoulli(\epsilon)$\;
        \If{e=1}{
        $a_t$ $\leftarrow$ sequentially select sparse arms\;
        }
        }
    \If{alert = red card}{
        $a_t$ $\leftarrow$ explore through $a_1$ to $a_k$\;
        $q_{\theta^P}$ $\leftarrow$ Initiator($L^P$, n)\;
        
        $p(\theta^P_i)$ $\leftarrow$ $L$, $p(\theta)$ $\leftarrow$ $\frac{p(\theta)}{\Sigma_{\theta}{p(\theta)}}$}
offer price $a_{t, k}$ and observe $d_{t, a_t}$\;
alert $\leftarrow$ transferability test($d_{t, a_t}$)\;
$p_{t+1}(\theta) \leftarrow \frac{p_t(\theta)q_{\theta, a_t}(d_{t, a_t})}{\Sigma_{\theta'\in\Theta}p_t(\theta')q_{\theta', a_t}(d_{t,a_t})}$
}

\end{algorithm}

\vspace{-3mm}
\subsection{Universe synthesizer}
\vspace{-3mm}
\label{subsec:env_synth}

To exploit the indirect information between available prices, we model the environment uncertainty via a set of universes $\theta$s. Each universe $\theta$ associates a distinctive customer valuation distribution and consequently a demand curve. Such association treats ''set of available prices`` as a fundamental modeling unit, instead of modeling the uncertainty of each individual action as \texttt{UCB} and \texttt{TS}. Using the bayesian language, we further defines the likelihood of observing $d \in 0,1,\dots,N,$ purchases when price $a$ is set under the $\theta$ universe as $p(d|\theta, a)$. With some abuse of notation, we denotes $p(d|\theta, a)$ as $q_{\theta, a}(d)$ and let $q$ be the likelihood of all possible outcomes corresponds to the proposed price as a whole. Mathematically, $q$ is a $L \times k \times (N+1)$ vector.

Universe synthesizer provides the likely bandit environments $\Theta$ = \{$\theta:1, 2, ..., L\}$ and $q_{\theta}$. $p(\theta)$ captures the belief about the true nature of the system. While $q_{\theta}$ do not update on the fly, $\Theta$ can be expanded and include more appropriate $q_{\theta}$. As a result, demand curve is learned by updating the belief on all universes, and the information obtained from calculating the posterior can be shared by all arms. Hence, ACIDP could transfer to a new environment agilely.

There are three source of bandit environments involved:  Perceived universes (environment exploration), Vintage universes (environment exploitation), and Counterfactual universes (environment experimentation). For simplicity, we denote ${\Theta} = {\Theta^{P}} \cup {\Theta^{V}} \cup {\Theta^{C}}$ and $L=L^{P}+L^{V}+L^{C}$.
\\ \hspace*{\fill} \\
\textbf{A.Perceived universes (environment exploration).}
In finite set IDS, $q$ is treated as a known input. Cases such as new product launch, however, do not guarantee $q$ to be known. Perceived universe generalize $q$ by \textit{learn-then-IDS}. Algorithm 2 samples $n$ times from each price $a$ and summarize the empirical likelihood as $q_{\theta^P}$. Number of repetitive sample $n$ and increment perceived universes $L^{P}$ are left as hyperparameters. This naive method consumes $L^P*K*N*n$ customer exploration. Particularly, $N*n$ are the sample size for an atomic estimation of $q_{\theta,a}$.
\begin{algorithm}
\caption{Initiator($L^P$, n)}
\SetAlgoLined
\DontPrintSemicolon
\For{i = 1 to $L^P$}{
    \For{1 to n}{
        \For{a = $a_1$ to $a_k$}{
        Sample $a$ and collect observation $d_{a, n}$\;
        }
    }
    $q_{\theta^{P_i}, a}(d)$ $\leftarrow$ $\frac{1}{n}\Sigma_n I(d_{a, n}=d)$ $\forall d, a$\;
}
Return $q_{\theta^P}$
\end{algorithm}

\textbf{Remark.} If $n>1$, we encourage first loop through $K$ then $n$ to avoid inconsistent estimation across $a_k$ within a single universe when non-stationarity presents. Besides, zero probability needs to be replaced with a tiny number for unobserved events for computational consideration.

\textbf{B.Vintage universes (environment exploitation).}
Given past data, or prior belief, we could prepare them as $q_{\theta^V}$ and $p(\theta^V)$ as discussed. Such source would then be checked by Bayesian updating or refined through counterfactual universes.

\textbf{C.Counterfactual universes (environment experimentation)}
Computing $q$ with pure Monte Carlo methods could reduce inevitable regret from exploration. Nonetheless, it may took unlimited guess and is computationally inefficient. Algorithm 3 implements such spirit by integrating previous universes to estimate $D(a)$ and $f_V$ to generate diverse universes $\theta^C$, $q_{\theta^C}$. Counterfactual universe are generally cost-friendly and could conspicuously extend the previous two sources.

\begin{algorithm}
\caption{Generator(c)}
\SetAlgoLined
\DontPrintSemicolon
$D(a) = \Sigma_{\theta}\Sigma_{d}p(\theta)p(d|\theta, a)d/N$, $\forall a$\;
$\bar{a_k} = \frac{a_k + a_{k+1}}{2} , k=1,2,\dot,k-1$\;
$p(V'=\bar{a_k})$ = $D(a_k) - D(a_{k+1})$, $k=1,2,\dot,k-1$\;
\For{$c_i$ $\in$ $[-a_k, a_k]$}{
$f_{V^{C_i}} \sim p(V')+c_i+\xi$, where $\xi \sim N(0,\sup\{\bar{a_k}\})$\;
\For{$j = 1\ to\ n$}{
    sample $v^{C_i}_j$ from $f_{V^{C_i}}$
    }
$q_{\theta^{C_i}, a}$ $\leftarrow$ $p(\Sigma^n_{j=1} I(v^{C_i}_j\geq a)=d)$, $\forall$ $d, a$\;
}
Return $q_{\theta^C}$
\end{algorithm}

In Algorithm 3 (visualized in appendix \ref{fig:vis_gen1}), line1 estimates the demand curve by calculating the posterior predictive distribution and normalize all components into $[0, 1]$ by dividing $N$. Line2 uses the midpoint of price intervals to estimate valuation distribution. Since we can only observe the realized demands that associates with the available price sets $\{a_k\}$, the domain of $d(a)$ is hence discrete and unlike the underlying valuation distribution $f_{V_t}$. Line2 uses the midpoint for a naive interpolation. Namely, say $D(0.5)=0.4$ and $D(0.6)=0.3$, we conclude that there are $10\%$ of customers holds a valuation near $0.55$. Line3 derives a discrete approximation of $f_{V_t}$ using midpoints $\{\bar{a_k}\}$. Line4 then decorates the final approximate $f_{V^C}$ by applying noise $\xi$ and intended manipulation $c$. The noise creates a smooth continuous texture and is responsible for neutralizing the sampling error with randomization. $c$ could be any known manipulation. Line 5 demonstrates a simple linear transformation within the price range. Line6 samples from the artificial valuation distribution $f_{V^{C_i}}$ to construct empirical likelihood $q_{\theta^{C_i}}$.

Overall, online samples, historical data and synthetic data are main resource of information. ACIDP collaborate with all of them and functions as an ensemble method. In the Bayesian inference phase, the belief is updated recursively and each source are assigned with a distinct weight by the posterior distribution to maintain credible bandit environment.

\textbf{Remark.} In practice, one should initialize higher prior on $\theta^P$ to reflect the authenticity. Besides, we encourage setting a tiny lower bound for $p(\theta)$ such that the universe wouldn't be completely ruled out by Bayesian update. \ref{sec:exp_to_imple} Case 4, 5 exemplify that old $\theta s$ are recyclable and could benefit decision-making.

\vspace{-3mm}
\subsection{Information-directed pricing}
\vspace{-3mm}
\label{subsec:InfoDir_pricer}
The information-directed pricing ensures the optimality by lowering expected regret and gaining information as much in a single period. While Beta-Bernoulli bandit is feasible for binary purchase result, we suggest that finite setting is more flexible and computationally convenient. The likelihood $q_{\theta}$ could be designed to capture specific relation in prices such as correlation and monotonicity, and it does not rely on parametric prior and likelihood distributions. IDP optimize the price by exploring the multi-universe $\Theta$. There are three steps covered in IDP.

\textbf{A. Estimate Regret and Information Gain}
Conditioned on universe $\theta$, the next observation $d_{t}(a_{t})$ given price $a_{t}$ is estimated by $q_{\theta, a_{t}}(d)$. The prior $p(\theta)$ denote the belief of the universe $\theta$. Let $\Theta_{a}$ denotes the collection of universes $\theta$ that the price $a$ is the optimal price where the optimal price $a$ is the price that maximize the expected profit $\sum_{d}q_{\theta, a}(d)r_{a}(d)$ of offering price $a$. Thus, the optimality of price $a^*$ can be described by $p(a^{*})=\sum_{\theta \in \Theta_{a^{*}}}p(\theta)$.
Here, we further define the expected regret by the gap from optimal aggregate profit of offering price a $R^{*}-\sum_{\theta} p(\theta) \sum_{d} q_{\theta, a}(d) r_a(d)$, where $R^*$ is the possible expected optimal reward.
Finally, we use the the mutual information $I(a^*, d)$ = $\sum_{a^{*}, d} p_{a}\left(a^{*}, d\right) \log \frac{p_{a}\left(a^{*}, d\right)}{p\left(a^{*}\right) p_{a}(d)}$ to measure the expected information gain. $p_{a}(d)$ is the aggregate demand distribution of price $a$ and $p_{a}(a^{*}, d)$ is the aggregate price-demand joint distribution inside $a^*$-universes. Algorithm 4 summarizes the complete process.
\begin{algorithm}
\label{alg:FiniteIR}
\caption{FiniteIR(L, K, N, p, q)}
\SetAlgoLined
\DontPrintSemicolon
$\Theta_{a}$ $\leftarrow$ $\{\theta|a = \arg \max_{a'}\Sigma_d q_{\theta, a'}(d)a' d\}, \forall \theta$\;
$p(a^*)$ $\leftarrow$ $\Sigma_{\theta \in \Theta_{a^*}} p(\theta), \forall a^*$\;
$p_a(d)$ $\leftarrow$ $\Sigma_{\theta}p(\theta)q_{\theta,a}(d),  \forall a,d,\theta$\;
$p_a(a^*,d)$ $\leftarrow$ $\frac{1}{p(a^*)}\Sigma_{\theta \in \Theta_{a^*}}q_{\theta,a}(d),\forall a,d,a^*$\;
$R^*$ $\leftarrow$ $\Sigma_a \Sigma_{\theta \in \Theta_{a^*}} \Sigma_d p(\theta)q_{\theta,a}(d)ad$\;
$\vec{g_k}$ $\leftarrow$  $\Sigma_{a^*,d}p_a(a^*,d)log\frac{p_a(a^*,d)}{p(a^*)p_a(d)}, \forall a$\;
$\vec{\Delta_k}$ $\leftarrow$ $R^* - \Sigma_\theta p(\theta) \Sigma_d q_{\theta,a}(d)ad, \forall a$\;
Return $\vec{\Delta}$, $\vec{g}$
\end{algorithm}

\textbf{B. Optimize with Information Ratio}
Intuitively, minimizing the square of expected regret divided by expected information gain balance between the reward and information trade-off. The information ratio is defined as
$\Psi_t(\pi) = \pi^{T}\Delta^2/\pi^{T}g$
where $\pi$ denotes the a k-dimensional action distribution vector and IDP then choose a price according to
$$\pi^{IDP} \in \arg\min_{\pi} (\pi^{T}\Delta)^2/\pi^{T}g$$
for monotone price sets, it is sufficent to pull the action with
$$a_k \in \arg\min_{k} \Delta_k^2/g_k$$

\textbf{C. Update universes belief}
Lastly, we update the belief on universes with new observation by
$$p_{t+1}(\theta) \leftarrow \frac{p_t(\theta)q_{\theta, a_t}(d_{t, A_t})}{\Sigma_{\theta'\in\Theta}p_t(\theta')q_{\theta', a_t}(d_{t,a_t})}$$
This step is vital for adaptation when confronted with environmental shift. While the goal of Information Ratio is to figure out the optimal price (ensure optimality), Bayesian update is more concerned with unveiling the true environment (maintain credibility). They interact with each other to settle on the ultimate solution.

\vspace{-3mm}
\subsection{Pricing strategy auditing}
\vspace{-3mm}
\label{subsec:pricing_strategy_auditing}
Stationary bandit algorithms struggle in non-stationary as their convergent decision blinded them from environmental changes. Interestingly, it forms another exploration-exploitation trade-off to check stationarity. Only when the present set of  $\theta s$ no longer applies should transferability test allow the ACIDP reinitialize its belief. Otherwise, manageable shifts should first be addressed by recruiting additional counterfactual universes and Bayesian update. ACIDP accomplish this by classifying the environmental changes into three states: no change, slight change, huge change. It is like 
there are two levels of alert in ACIDP for non-stationarity: Yellow Card and Red Card. The alert and the accompanying measure are specified as the three steps below. Algorithm 5 summarizes the framework. The first step start from line1 to line7; second at line8; the rest for final step. Here $w$ is a hyperparameter representing the maximum number of lastest periods considered contemporary, and the newest decision will based on observation found in the $w$-period rolling window.

\begin{algorithm}
\caption{Transferability Test($d_{t, a_t}$)}
\SetAlgoLined
\DontPrintSemicolon
$W_{\tau}$ $\leftarrow$ $\{d_{\tau, a_{\tau}}: d_{\tau, a_{\tau}} \in H_{\tau}, t\geq \tau \geq t-w\}$\;
$W_{\tau, a_t}$ $\leftarrow$ $\{d_{\tau, a_t}:d_{ \tau,a_{\tau}=a_t}, d_{\tau} \in W | a_t\}$\;
$\{d_i\}^m_{i=1}$ $\leftarrow$ reindex $W_{\tau, a_t}$\;
$\bar{d}$ $\leftarrow$ $\frac{\Sigma^m_{i=m-n}d_i}{n}$, $X$ $\leftarrow$ $\Sigma^{m-n}_{i=1} \frac{d_i-\bar{d}}{N/2}$\;
$LB$, $UB$ $\leftarrow$ $\frac{X}{m-n}$ $\pm$ $1.7$ $\sqrt{\frac{loglog(2t)+0.72log(10.4/\alpha_1)}{t}}$\;
\If{ $LB>0$ or $UB<0$ }{
    alert $\leftarrow$ yellow card\;
    \If{e=1 (follows from line7 in algorithm1)}{
        $pvalue = binomial(d_{t, a_t}, N, \Sigma_{\theta} p(\theta)p_{\theta, a_t}(d)d/N)$\;
        \eIf{$pvalue$ $<$ $\alpha_2/n$}{
            alert $\leftarrow$ red card
        }{$\epsilon$ $\leftarrow$ $\epsilon * r_{decay}$}
    }
}
Return alert
\end{algorithm}

\textbf{A. Martingale Test for stationarity. (Yellow card: Demand model credibility inspection)}
In each round, the martingale sequence test compares the empirical mean of most recent $n$  $d_{\tau, a_\tau = a_k}$ with the earlier $m-n$ within $w$ periods with confidence level $\alpha_1$. Line1 gives the latest $w$ period data from filtration $H_{\tau}$, from which we filter out $m$ of those associated with same price as $a_t$. Line4 reallocate the observation into latest $n$ and not as latest $m-n$. $X$ is the test statistic measuring the deviation of the two ordered sequence, and 2 in denominator is the recommended adjustment for binomial distribution according to Howard. Line5 then move on to test the deviation of the two sequence, $LB$ denotes lower bound, $UB$ denotes upper bound (equation 2 in \citep{howard2021time}). A yellow card alarms ACIDP when anomaly detected. Yellow Card first summons Generator for possible environment changes. Besides, an $\epsilon$-auditor is on standby to sample and execute diagnostic inspection with probability $\epsilon$.\

\textbf{B. $\epsilon-$auditor sampling (Verification: Price mechanism optimality inspection)}
With $\epsilon$, $\epsilon-auditor$ explores the unfamiliar arms and collects evidence for non-stationarity; otherwise, IDP keeps exploiting on what it has. For diagnostic tests, it is usually sufficient to sample the quartiles over the price range. So the regret do not necessarily grow with K. In practice, the auditor is activated after $p(\theta)$ converged to ensure the upfront change is unpredictable and $D(a)$ has stopped improving from existing knowledge. The $\epsilon$ decays by $r_{decay}$ when the diagnostic tests are not rejected, and resets whenever receives another yellow card.

\textbf{C. Demand Deviation test. (Red card: Optimal price transferability inspection)}
Here, one could adopt correspondent hypothesis testing techniques that suits the observation structure. In our setting we conduct binomial tests with $binomial(d_{t, a_k}, N, \Sigma_{\theta} p(\theta)p_{\theta, a_k}(d)d/N))$ on n prices collected by $\epsilon-$auditor with Bonferroni corrected confidence level $\alpha_2/n$. If the test is rejected in arbitrary price, we conclude that the current demand belief is obsolete. This may indicates both external or internal errors, i.e. 1. the environment has changed, or 2. a larger sample size is required for credibility and optimality. In either way, the Initiator will be triggered and facilitates exploration.

\vspace{-5mm}

\vspace{-1mm}
\section{Experiments}\label{sec:exp_to_imple}
\vspace{-2mm}

This section discusses the computational findings of our proposed method. We highlighted simulations that covers six distinct common situations and an real-life example with corporate data. We begin by simulating the well-studied stationary MAB pricing scenario. On to some non-stationary bandit settings with structural shifts, four scenarios are examined: rapid growth, rapid decline, seasonality, and volatility. Next, we consider an upside-down move in market environment where we suppose the shift is fully unanticipated. Finally, we implements ACIDP in real customer data.

We adapt the experiment design from \citep{misra2019dynamic}. The following data generation processes are identical in six simulation. Customers arrive, notice the price, and only purchase if their valuation $v_{i,s}$ exceeds the price. All policies revise the pricing every N=10 customers, so $d_t \in {0, 1,.., 10}$ purchases. Customers are assumed to come from 1000 distinct equal-sized segments that holds distinct valuation $\{v_s\}^{1000}_{s=1} \overset{i.i.d}{\sim} Beta(3, 6)$ which is a right-skewed distribution. Besides, we consider heterogeneity within the segments by $v_{i,s} = v_s + e_i, e_i \sim N(0, 0.1)$. We consider this setting as the beginning customer valuation $f_{V_0}$, and valuation distribution $f_{V_t}$ will evolve from this baseline as time goes (except Case 3).

All policies are unaware of the $f_V$, the shifting moments and magnitude. Furthermore, $v$ is not directly observable. The agent must learn the demand curve under partial monitoring and choose from the price set $p \in [0.01, 1]$ with $K = 20$ prices evenly spaced to minimize regret. All cases are subjected to a $T=2000$ horizon and ten independent repetitive trials. Along with ACIDP, we also implement UCB family, TS and EG, policies that do not require T as known. We have attempted to delineate the implemented policies are optimized for the simulated horizon.

\begin{figure*}
  \centering
  \includegraphics[width=1\linewidth,page=3]{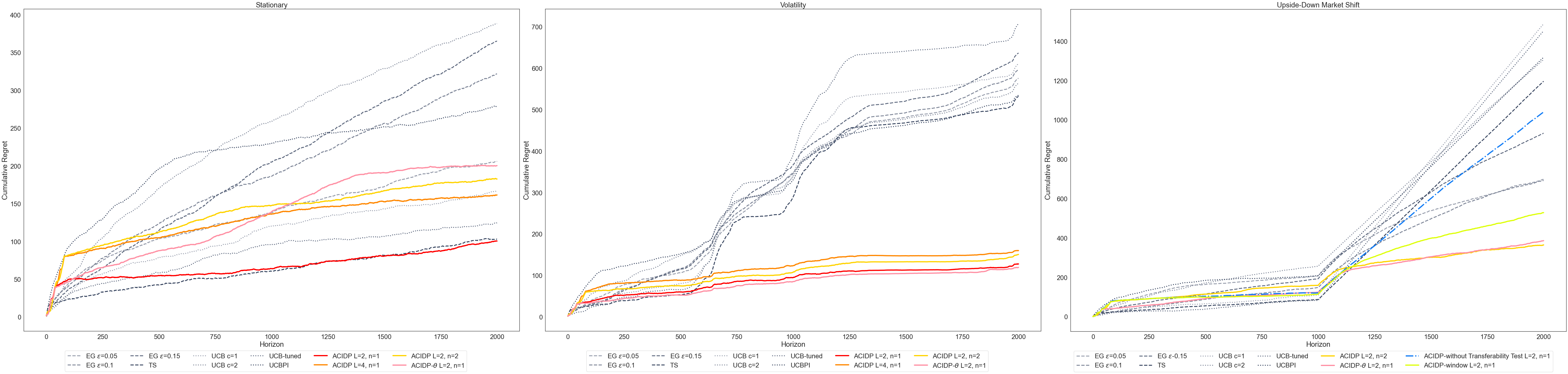}
  \caption{Mean Regret Under Stationary, Structural Non-stationary, Upside-Down Non-stationary .}\label{fig:case1regret}
\end{figure*}

\begin{figure*}
  \centering
  \includegraphics[width=1\linewidth,page=3]{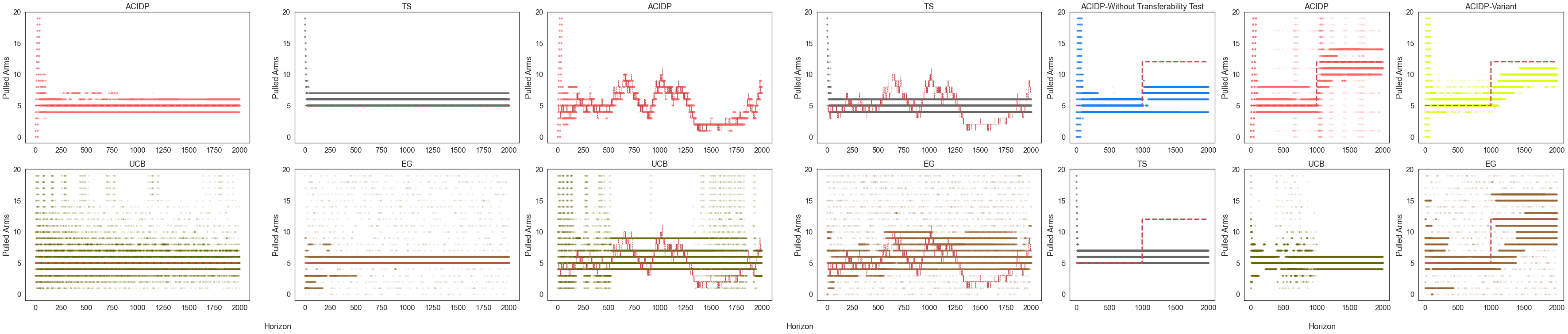}
  \caption{Price Selection Sequence Under Stationary, Predictable Non-stationary, Unpredictable Non-stationary (All Trials)}\label{fig:case1arm}
\end{figure*}

\vspace{-3mm}
\subsection{Stationary environment}
\vspace{-3mm}

\textbf{Case1: Stationary Environment}
Under stationary conditions, candidates appear to explore adequately and are capable of recognizing the true optimal price. Note that ACIDP-$\theta$ has a somewhat greater mean regret when L=2 and n=1. It result from observing outliers upon calculating empirical mean in perceived universes in a single trial, which also explains why IDS with L=4, n=1 yields a higher mean and perform robustly across trials. In general, we demonstrated that, despite accruing fixed initialization cost, learn-then-IDS is still competitive with various alternative policies in stationary bandit settings.

\vspace{-3mm}
\subsection{Structural Environment Shifts}
\vspace{-3mm}

\textbf{Case2: Rapid Growth.}
In the second case, the $V$ distribution experience a rapid growth of 0.3 at t=1001. i.e. $V_1, V_2, \dots, V_{1000} \sim f_{V_0}$, and $V_{1001}, V_{1002}, \dots, V_{2000} = V_{1000}+0.3$. This setup can be compared to the effect of governmental revitalization, in which all customers received the same level of subsidy.

When confronted with upward adjustments, TS and UCB typically cease exploring immediately and converge on $a_{1000}$ because they believe that their choice has started to proper even more without comparing the growth in other options. Thus, their regret grows linearly after t=1000. As for ACIDP, this slight change is easily detected by the martingale tests
and resolved by the counterfactual universes.

\textbf{Case3: Rapid Decline.}
The third case is the reverse counterpart of case2. It starts at $V_1 = V_0 + 0.3, V_0 \sim f_{V_0}$ and the valuation drops to $V_{1001}, V_{1002}, \dots, V_{2000} \sim f_{V_0}$. The purpose of adopting such design is to allow clear comparison with case2.

Though the expected cumulative optimal rewards are identical in Case2 and Case3, the performance of UCB family plummets in Case3. It shows clear defect in the optimistic decision. Optimistic decision becomes determinant and overconfident when facing upwards trends while being indecisive when facing downwards shifts. Nevertheless, ACIDP excels in both cases.

\textbf{Case4: Seasonality.}
The fourth case considers the presence of a periodic shift such that $V_t= V_0 + 0.3\sin(4t\pi/T), V_0 \sim f_{V_0}$. The purpose of this example is to simulate the seasonality in customer preferences or market trends.

This case showcases the pivotal role of generator and IDP. Alternatives such as epsilon-greedy may be able to reinvigorate to sudden structural changes thanks to randomized checks, these checks are too inefficient to recover from recursive or continuous changes. In a periodic non-stationary environment, the perceived total environmental shifts offset over time, rendering alternatives ineffective. Alternatives relies on a single global mean estimation, so the mean returns to the bench mark after a cycle of environmental changes. ACIDP, on the other hand, discovers the optimal arm by first recording each presence in universes. Appeared universes could then be restored and recycled, which explains why sin function curve (shown in appendix \ref{fig:case4}) becomes more defined in the later phase.

\textbf{Case5: Volatility.}
In the fifth case, we consider $f_V$ with volatile structural shifts dependent on t. We take an one-dimensional Brownian motion $\{B_t\}_{T\geq t \geq1}$ and shift $V_t = V_0 + B_t$ by $B_t = B_{t-1} + \xi/\sqrt{T}$, where $\xi \overset{i.i.d.}\sim N(0, 1)$. Given the independence of the movements in each round, we postulated that the safest strategy to minimize regret is to price at $a^*_{t-1}$ $\forall t>1$ as $\mathbb{E}[B_t] = B_{t-1}$. As a result, the policy must respond quickly and robustly in light of the filtration.

ACIDP adapts gracefully to the volatile environment. It is sensitive to minute changes and respond precisely. Again, this volatile example is associated with a possible shift offset. Specifically, the range of $V$ experiences a similar peak at around the $700^{th}$, $1050^{th}$, and $2000^{th}$ period in this sequence. Not only did ACIDP modify its prices to reflect the fluctuation throughout the stochastic sequence, it maintained constant during these upswings. Such robustness is a desirable characteristic in dynamic pricing.

\vspace{-5mm}
\subsection{Upside-down Environment shift}
\vspace{-3mm}

\textbf{Case6: Distribution Shift.}
In the last synthetic scenario, we investigate an extreme example in which the hidden value distribution changes unexpectedly. At t=1000, the population distribution of segment mean set ${v_s}^1000_{s=1}$ transforms from a right-skewed $beta(3, 6)$ to a bimodal $beta(0.9, 0.5)$ with a greater mean and variance. We also let the mean difference between the two distributions to be similar as case 2.

After $1000^{th}$ round, only ACIDP have identified the true optimal arm; other policies' regrets grow linearly with time. Additionally, we establish that, while ACIDP is capable of acting in near-optimal directions without transferability test, it does not recognize the global optimal arm. The regular ACIDP with transferability test learns the new optimal arm through the recruitment of new universes, the last resort of transferability test, whereas the ACIDP-window proposed in appendix \ref{alter} progressively transitions to the new optimal through online fine-tuning its existing universes.

\subsection{Real-Life Data}
Finally, we test our algorithm in a realistic scenario with Criteo Sponsored Search Conversion Log Dataset \citep{tallis2018reacting}. Criteo is a French online advertising firm. The data are collected from August to November 2018 and covers price, ad clicks, and conversion records for several products. We consider a non-stationary bandit pricing problem where during the pricing horizon, the e-commerce platform mistook the product information of A with another product B on the webpages, and the agents are required to fix the price tag automatically to avoid potential loss.\\
To begin, we selected three of the most popular products across a range of pricing points, and were able to model the demand curve of them based on the real data. From low to high price levels, we have product A, B, C with optimal prices €70, €150, and €280 respectively.\\
The experiment lasted for T=6000 rounds in total; initially, the agents were asked to price product B based solely on conversion rate. At t=2000, the page was mistaken with product C and thus the original price tag seemed too appealing. The regret may increases dramatically if the price was not adjusted accordingly. Similarly, at t=4000, the page was again mislabeled with information of product A. In this case, the prices are expected to go from €150 to €280 then €70.\\
The action set consists of ${a_k}^50_{k=1}$ = ${10, 20, ..., 500}$ $K=50$ prices, which are the price levels most products collaborated with Criteo at. The price reset every $N = 500$ clicks. The only information available each round is the number of clicks that contribute to conversion. The policies are tested with noisy conversions generated from $d_{t,a_t} \sim binomial(N=500, D_{t,a_t})$, where $D_{t,a_t}$ is the true conversion rate calculated from the dataset.\\
The result is align with the above findings. It showed that ACIDP is the only policy capable of achieving optimality and pricing all three products rapidly and accurately. The cumulative regret of ACIDP exhibited sub-linearity and id significantly lower comparing to others. While most of the policies are capable of recognizing the optimal price of product B, ACIDP's success in detecting the inability to transfer existing knowledge to price product A, C at around $2000^{th}$, $4000^{th}$ round and activating the initiator to retrain itself is what makes the transparent difference.
\begin{figure*}
  \centering
  \includegraphics[width=1\linewidth,page=3]{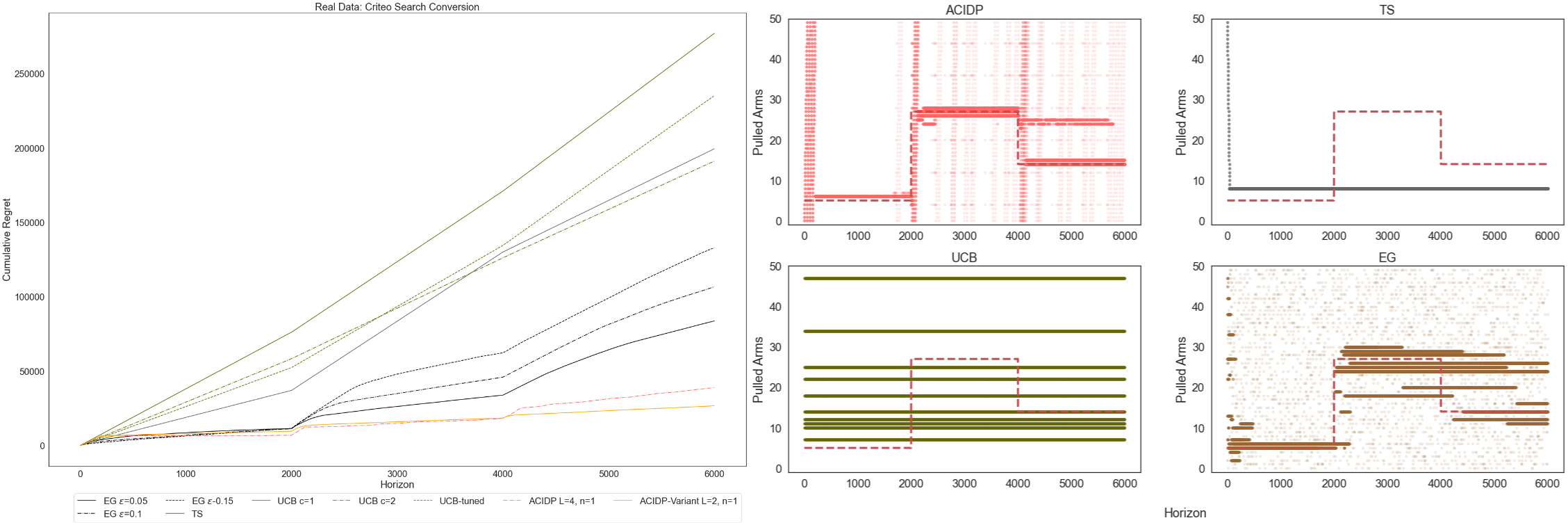}
  \caption{Regret and Price Selection (red line denotes $a^*$) in Real World Data.}\label{fig:criteo}
\end{figure*}

\vspace{-5mm}
\section{Conclusion and Future works}
\vspace{-3mm}
This article explores dynamic pricing in MAB with non-stationary rewards and introduces a new algorithm, Actor-Critic Information-Directed Pricing. The proposed method takes into account the presence external changes and exploits regret-minimized prices while identifying market demand on the fly. We investigate qualities that are absent in classical solutions concerning dynamic pricing problems and address them by developing three core modules: Universe Synthesizer, Transferability Test, and Information-Directed Pricing. Universe Synthesizer deals with price correlation by assessing the common source of demand and simulating possible futures in parallel universes. The transferability test decomposes the non-stationarity in the environment into three steps. First, Universe Synthesizer collaborates with IDP to deal with controllable changes. A step further, ACIDP undergoes thorough exploration once the current knowledge had been concluded outmoded. IDP embraces ensemble and information-directed sampling techniques to estimate the best price from the information presented in sources offered by the Universe Synthesizer. These modules are empirically proved to outperform alternative policies that are typically dominated by exploiting the empirical means in separate arms.

Questions remain unanswered. To begin, whether one could more efficiently initiate $q$ in dependent arm scenarios. Secondly, as we aim to price according to customer valuation, it's worth researching whether ACIDP works effectively when competition exists.  Finally, one could further consider heterogeneous variance across different segments. This may lead to a non-uniform estimation error across the valuation.

\clearpage

\nocite{*}
\bibliography{main}
\clearpage

\begin{appendices}

\section{Alternative Design}\label{alter}
\subsection{Variant of \texttt{ACIDP}}
\textbf{Moving-window Transfer}
An alternative design of a transferable bandit algorithm assumes the newest the best. In algorithm 6, we present a method that evolves without hypothesis testing and additional exploration. This algorithm works well when each shifted stage lasts for a sufficient amount of time. Let $\theta^w$ be a specialized counterfactual universe that updates itself constantly. $\theta^w$ learn from others' experience and overwrites recent pulled price in  $q_{\theta^w}$ with latest samples. Since such manipulation may violate the law of non-increasing demand over ordered prices. So in line 9, assuming $a_k < a_{k+1}, \forall k$, we correct it by the estimates of adjacent prices. The recorded likelihood will be inherited proportion to the posterior $p(\theta^w)$, i.e. if such update is indeed plausible, the $\theta^w$ will dominates $\Theta$ and reproduce itself in the next round.
Ideally, ACIDP-window selects a monotone pricing sequence till they reach the near-optimal arms. The hyperparameter $W$ controls the transfer speeds in this case. This approach sometimes outperforms ACIDP as it does not require additional exploration. (Shown in Simulation Case 6) Intuitively, the algorithm updates the likelihood by assuming recent samples are truth and normalization gradually dilute the weight of older samples and transfer $q_{\theta^w}$ into the new environment.

\begin{algorithm}
\caption{Transferability Test-Window Variant(w)}
\SetAlgoLined
\DontPrintSemicolon
$W_{\tau}$ $\leftarrow$ $\{a_{\tau}, d_{\tau, a_{\tau}}: a_{\tau}, d_{\tau, a_{\tau}} \in H_{\tau}, t>\tau \geq t-w\}$\;
\While{$T > t > w$}{
$q_{\theta^w}$ $\leftarrow$ $\Sigma_{\theta}p(\theta)p(d|\theta)$\;
\For{$a_k \in W$}{
    $q_{\theta^{W}, a}(d)$ $\leftarrow$ $\frac{|\{d_{a_k}: d_{a_k}=d, d_a \in W|d\}|}{|\{a_k: a_k \in W\}|}$ $\forall d$\;
}
$D^w(a)$ $\leftarrow$ $\Sigma_d q_{\theta^w, a}(d)d/N$ $\forall a$\;
\eIf{$d_{t, a_t}>\Sigma_{\theta}\Sigma_d p(\theta)p(d|\theta, a_t)d$}
{$D^w(a_k)$ $\leftarrow$ $max(D^w(a_k), D^w(a_{k-1}))$, $\forall k$}
{$D^w(a_k)$ $\leftarrow$ $min(D^w(a_k), D^w(a_{k-1}))$, $\forall k$}
$q_{\theta^w_a}(d)$ $\leftarrow$ $binomial(d, N, D^w(a))$, $\forall d, a$
}
\end{algorithm}

\textbf{Information Structure}
In \citep{russo2018learning}, they looked at how the algorithm invests in indirect information when $\vec{g}$ is approximated by $I(\theta; d)$ instead of $I(a^*; d)$. In our setting, $a^*$ are geared with with $\theta$, i.e. in universes where individuals are willing to pay more, the optimal prices are also higher.Thus, the variant ACIDP-$\theta$ helps reduce the uncertainty about $\theta$ and improve demand curve estimation, which follows by the trivial solution of optimal price.

\subsection{Other Policies}
\textbf{$\epsilon$-Greedy (EG)}
At each time t, the algorithm choose the price that yields the highest empirical reward with $\epsilon$ $\in (0,1)$ and randomly pick the other $k-1$ prices with $1-\epsilon$.

\textbf{Upper Confidence Bound (UCB)}
UCB family are well-developed in bandit literatures. UCB make optimistic decisions by selecting the price with the highest reward upper bound $\bar{r_a} + c\sqrt{\log t/n(a_k)}$ where $n(a_k)$ is the number of times $a_k$ is pulled and $c$ is a hyper-parameter that motivates exploration. UCB-tuned substitutes c with $min(\frac{1}{4}, V)$. This tunable version takes the variance in outcomes into account where $V=\frac{1}{n} \sum r(a)^2 - \bar{r(a)^2} + \sqrt{2 \log t/n}$.

UCBPI uses $argmax\ \bar{r_a} + a_k \sqrt{\log t/n(a_k)}$ along with a rule-based partial identification technique to filter out the obvious sub-optimal prices initially. This version was built specifically for dynamic pricing in stationary environments and empirically encourage exploration in higher prices.

\textbf{Thompson Sampling(TS)}
Thompson sampling choose according to the posterior probability that each arm is perceived as the best 
$p(a_t = a | H_{\tau}) = p(a^*=a | H_{\tau})$
where $a^*= argmax\ \mathbb{E}(r(d)|\theta)$.

\section{Simulation}
In this section we discuss the details and additional findings in our simulation. Other parameters used in the simulation setting: $\epsilon=0.1$, $r_{decay}=0.1$, $\alpha_1=0.05$, $\alpha_2=0.01$, $w=300$, $n=5$ in sequence test. The performance are summarized in the following sections.

\subsection{Case1: Stationary}
\begin{figure*}[h]
  \centering
  \includegraphics[width=1\linewidth,page=3]{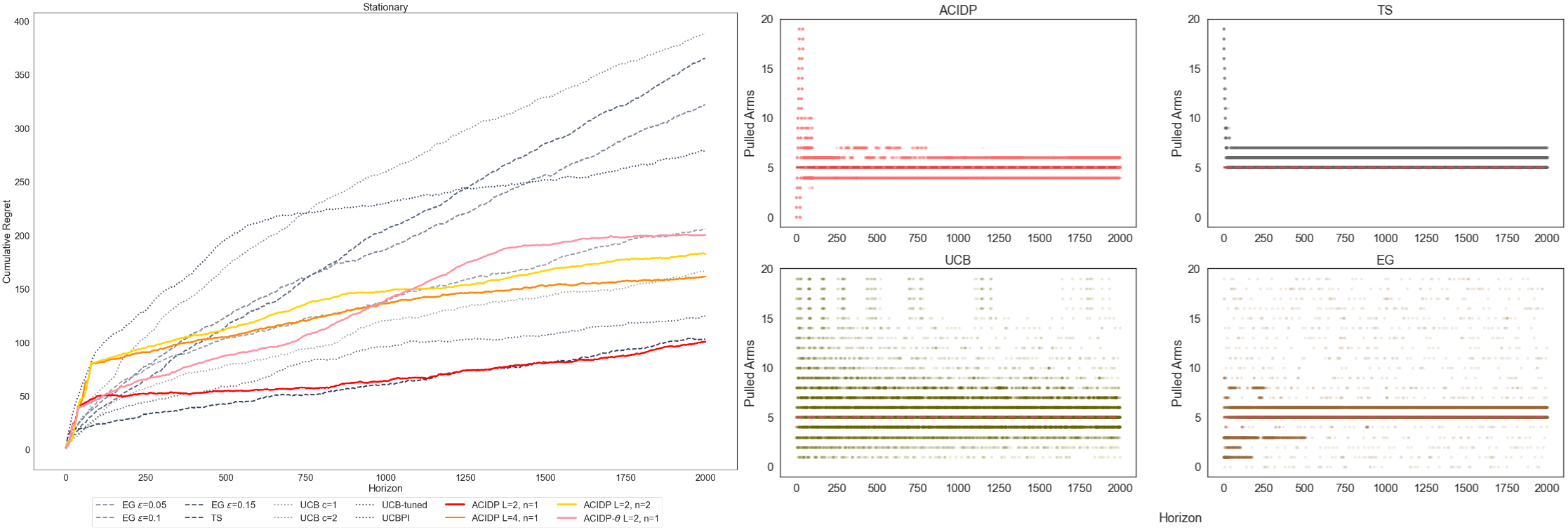}
  \caption{Regret and Price Selection (red line denotes $a^*$) in Case1: Stationary.}\label{fig:case1}
\end{figure*}

In the stationary case, all candidates perform satisfactorily. To demonstrate the impact of sample sizes on exploration, we utilize the ACIDP with hyperparameters {L=2, n=1}, {L=4, n=1}, {L=2, n=2}. Both of the latter two cost 80 rounds at initialization, so their regret develops more steeply at the start (see figure \ref{fig:case1}). They do, however, promptly detect the optimal arms and have minimal regret for the remainder of the time. ACIDP-$\theta$, on the other hand, performs relatively well in previous periods but suffers from an insufficient investigation in a single trial.

\begin{table}[h!]
\centering
\caption{Case1 Result}
    \begin{tabular}{ll|llll}
    \hline
        Policy & Hyperparameters & Mean Regret & Standard Error & Max & Min \\ \hline
        ACIDP & L=2, n=1 & 100.5 & 56.4 & 238.5 & 53.3 \\ 
        ~ & L=4, n=1 & 161.5 & 93.7 & 339.4 & 84.4 \\ 
        ~ & L=2, n=2 & 182.6 & 89.9 & 393.5 & 78.1 \\ 
        ACIDP-$\theta$ & L=2, n=1 & 200.3 & 243.0 & 867.4 & 41.2 \\ 
        EG & $\epsilon$=0.05 & 206.2 & 81.9 & 405.9 & 125.4 \\ 
        ~ & $\epsilon$=0.1 & 321.8 & 99.9 & 493.9 & 229.5 \\
        ~ & $\epsilon$=0.15 & 365.1 & 70.7 & 511.8 & 286.0 \\ 
        TS & ~ & 102.9 & 108.8 & 395.9 & 9.3 \\ 
        UCBPI & ~ & 278.9 & 26.4 & 319.3 & 230.8 \\ 
        UCB-Tuned & ~ & 124.5 & 19.1 & 160.9 & 95.9 \\ 
        UCB & c=1 & 165.7 & 8.5 & 182.1 & 153.5 \\ 
        ~ & c=2 & 388.3 & 28.6 & 436.5 & 344.7 \\ \hline
    \end{tabular}
\end{table}

\clearpage

\subsection{Case2: Rapid Growth}
In the case of rapid growth, policies are essential to achieve positive demand growth. For all trials, UCB and TS maintain the convergent decision at the $1000th$ round. Because their inertial decision and increased observation causing them to consider the convergence price is extraordinary at $1000^{th}$. Aside from ACIDP, EG is the only candidate that transfers to a somewhat superior yet optimal arm on occasion. This is understandable since it is less familiar with higher prices at $1000^{th}$, i.e. smaller sample sizes, and so the empirical mean is easily inflated, leading to choice modifications. In this case, ACIDP succeeded by detecting the change and inviting the generator to mimic a bandit environment to transfer to.

\begin{table}[h!]
\caption{Case2 Result.}
    \centering
    \begin{tabular}{ll|llll}
    \hline
        Policy & Hyperparameters & Mean Regret & Standard Error & Max & Min \\ \hline
        ACIDP & L=2, n=1 & 157.3 & 170.2 & 619.4 & 33.4 \\ \
        ~ & L=4, n=1 & 167.5 & 91.5 & 406.7 & 91.0 \\ 
        ~ & L=2, n=2 & 152.7 & 56.0 & 235.8 & 78.2 \\ 
        ACIDP-$\theta$ & L=2, n=1 & 135.0 & 107.2 & 379.0 & 50.9 \\ 
        EG & $\epsilon$=0.05 & 546.7 & 305.1 & 1224.3 & 219.3 \\ 
        ~ & $\epsilon$=0.1 & 759.0 & 389.2 & 1379.8 & 381.4 \\ 
        ~ & $\epsilon$=0.15 & 929.9 & 311.4 & 1484.0 & 565.7 \\ 
        TS & ~ & 766.7 & 226.5 & 1164.1 & 479.3 \\ 
        UCBPI & ~ & 900.7 & 15.9 & 922.4 & 872.0 \\ 
        UCB-Tuned & ~ & 1138.7 & 193.7 & 1248.1 & 741.2 \\ 
        UCB & c=1 & 1299.4 & 354.6 & 1752.8 & 778.8 \\ 
        ~ & c=2 & 1221.9 & 375.2 & 1894.8 & 576.3 \\ \hline
    \end{tabular}
\end{table}

\begin{figure*}[h]
  \centering
  \includegraphics[width=1\linewidth,page=3]{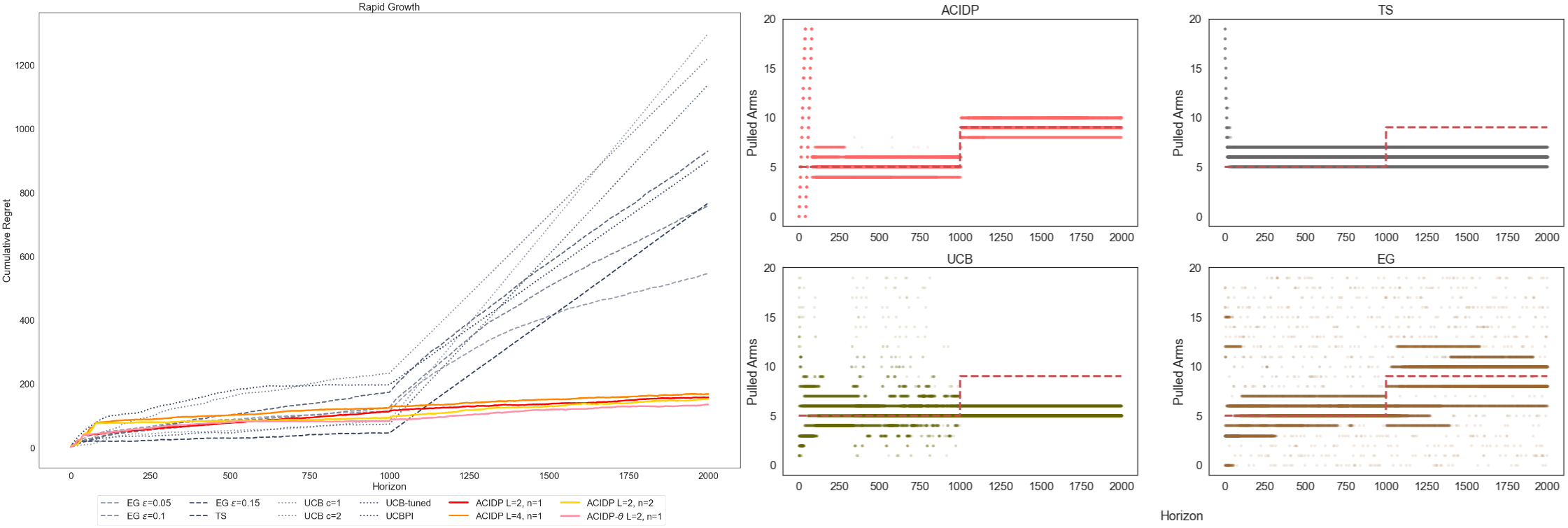}
  \caption{Regret and Price Selection (red line denotes $a^*$) in Case2: Rapid Growth.}\label{fig:case2}
\end{figure*}

\clearpage

\subsection{Case3: Rapid Decline}
In the case of rapid decline, candidates must implement price reductions in order to receive bigger rewards. The maximum cumulative rewards at the $2000^{th}$ round are the same as in the Rapid Growth case. Interestingly, TS, UCB, and EG behave radically differently and result in higher total regret in general. In contrast to the previous case, these policies lose faith in the previously convergent conclusion and rush to figure out the new one. However, due to prior experience, it takes more effort to investigate and identify the ideal pricing in the new environment. ACIDP, on the other hand, went through similar steps as the previous example and ended up with equivalent regret.

\begin{table}[h]
\caption{Case3 Result.}
    \centering
    \begin{tabular}{ll|llll}
    \hline
        Policy & Hyperparameters & Mean Regret & Standard Error & Max & Min \\ \hline
        ACIDP & L=2, n=1 & 194.7 & 120.9 & 399.0 & 53.2 \\ 
        ~ & L=4, n=1 & 206.7 & 88.3 & 416.2 & 112.7 \\ 
        ~ & L=2, n=2 & 171.8 & 42.1 & 228.8 & 94.3 \\ 
        ACIDP-$\theta$ & L=2, n=1 & 145.8 & 136.8 & 397.5 & 25.4 \\ 
        EG & $\epsilon$=0.05 & 869.9 & 243.4 & 1406.4 & 613.6 \\ 
        ~ & $\epsilon$=0.1 & 890.5 & 91.0 & 1114.8 & 800.1 \\ 
        ~ & $\epsilon$=0.15 & 952.5 & 71.1 & 1060.9 & 831.2 \\ 
        TS & ~ & 1620.1 & 347.3 & 2096.2 & 953.7 \\ 
        UCBPI & ~ & 1775.8 & 623.5 & 2158.6 & 572.4 \\ 
        UCB-Tuned & ~ & 1106.4 & 540.0 & 1782.4 & 445.5 \\ 
        UCB & c=1 & 1255.2 & 559.0 & 1911.5 & 376.9 \\ 
        ~ & c=2 & 742.8 & 26.5 & 793.5 & 708.2 \\ \hline
    \end{tabular}
\end{table}

\begin{figure*}[h]
  \centering
  \includegraphics[width=1\linewidth,page=3]{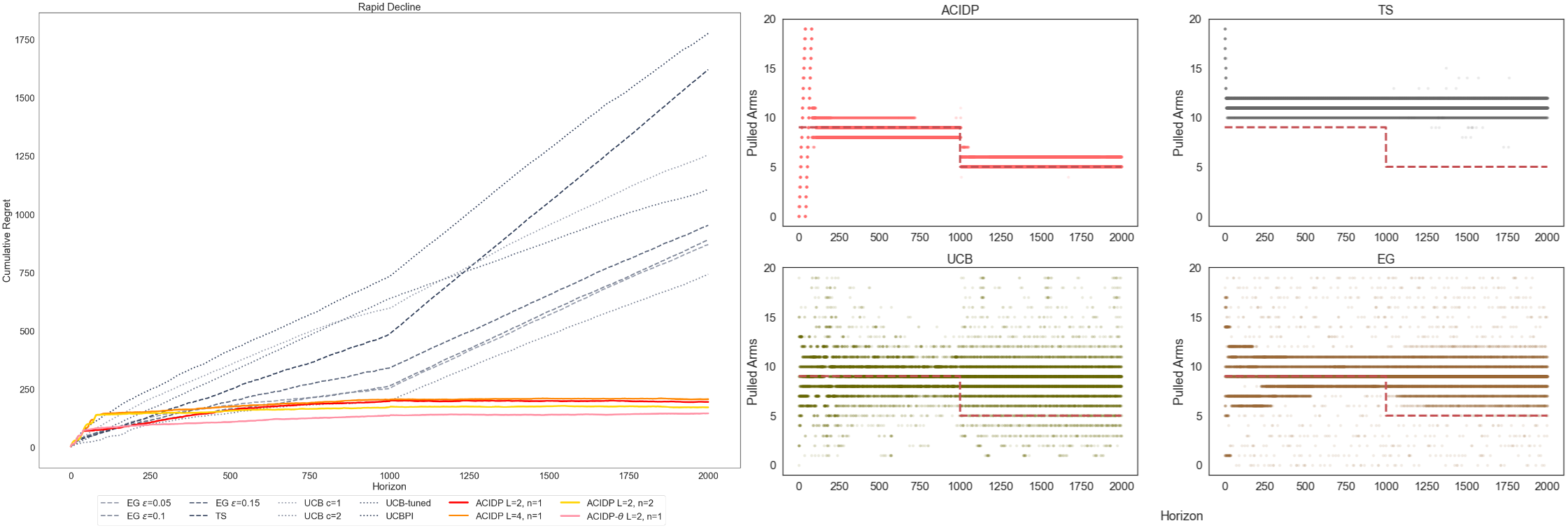}
  \caption{Regret and Price Selection (red line denotes $a^*$) in Case3: Rapid Decline.}\label{fig:case3}
\end{figure*}

\clearpage

\subsection{Case4: Seasonality}
In the case of seasonality, candidates need to overcome (1) continuous (2) simultaneous up and down shifts and may receive a bonus if they notice the recurrent change. In this case, UCB suffers from a combination of the two modes of confusion mentioned above, while TS continues to exploit at the same price. Because the observation is not identically distributed in each round, it is more difficult for an agent to effectively recognize the genuine optimal price in this case. However, in the later rounds, ACIDP can function optimally and leverage the previous experience..

\begin{table}[h]
\caption{Case4 Result.}
    \centering
    \begin{tabular}{ll|llll}
    \hline
        Policy & Hyperparameters & Mean Regret & Standard Error & Max & Min \\ \hline
        ACIDP & L=2, n=1 & 203.2 & 113.6 & 422.3 & 87.8 \\ 
        ~ & L=4, n=1 & 245.4 & 99.4 & 434.1 & 157.6 \\ 
        ~ & L=2, n=2 & 231.0 & 63.5 & 317.4 & 124.1 \\ 
        ACIDP-$\theta$ & L=2, n=1 & 260.0 & 83.6 & 437.3 & 188.9 \\ 
        EG & $\epsilon$=0.05 & 767.3 & 185.4 & 1217.3 & 553.2 \\ 
        ~ & $\epsilon$=0.1 & 761.4 & 149.2 & 1081.6 & 569.4 \\ 
        ~ & $\epsilon$=0.15 & 864.2 & 141.0 & 1149.0 & 686.5 \\ 
        TS & ~ & 481.5 & 23.1 & 500.1 & 424.3 \\ 
        UCBPI & ~ & 654.6 & 83.3 & 801.7 & 504.6 \\ 
        UCB-Tuned & ~ & 746.7 & 352.1 & 1449.5 & 408.2 \\ 
        UCB & c=1 & 580.0 & 110.0 & 787.8 & 439.9 \\ 
        ~ & c=2 & 657.4 & 117.0 & 945.0 & 517.3 \\ \hline
    \end{tabular}
\end{table}

\begin{figure*}[h]
  \centering
  \includegraphics[width=1\linewidth,page=3]{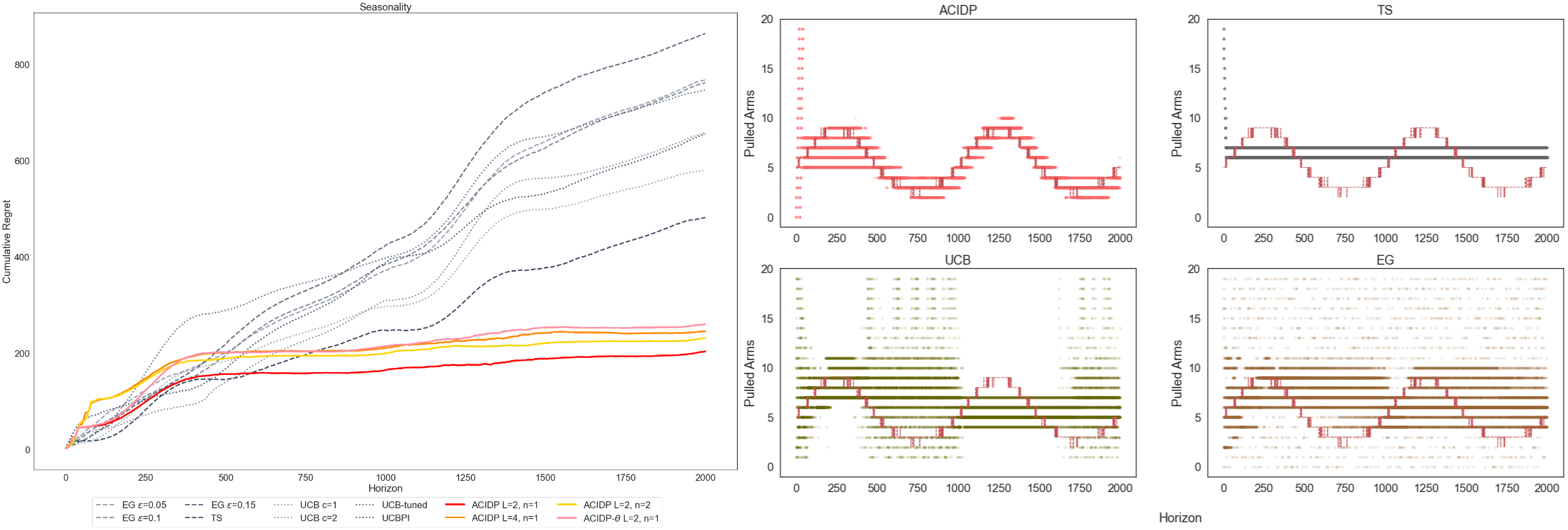}
  \caption{Regret and Price Selection (red line denotes $a^*$) in Case4: Seasonality.}\label{fig:case4}
\end{figure*}

\clearpage

\subsection{Case5: Volatility}
In the volatility case, the fluctuation is assumed to be independent from one period to another, observing the true optimal arm in the next round is impossible. However, we show that regret can be reduced by adapting rapidly enough and following the trend. We also discover that alternative candidates may price differently across the horizon under comparable demand levels. This is due to an inability to let go of past experiences, implying their performance can be considerably improved by learning when to forget past knowledge. The environmental diversity, on the other hand, is recorded as parallel universes in ACIDP; hence, obsolete knowledge does not constrain future decisions.

\begin{table}[h]
\caption{Case5 Result.}
    \centering
    \begin{tabular}{ll|llll}
    \hline
        Policy & Hyperparameters & Mean Regret & Standard Error & Max & Min \\ \hline
        ACIDP & L=2, n=1 & 127.5 & 25.8 & 164.8 & 93.1 \\ 
        ~ & L=4, n=1 & 159.8 & 28.3 & 221.3 & 124.9 \\ 
        ~ & L=2, n=2 & 150.4 & 30.3 & 190.8 & 100.5 \\ 
        ACIDP-$\theta$ & L=2, n=1 & 119.1 & 17.6 & 152.6 & 90.7 \\ 
        EG & $\epsilon$=0.05 & 575.9 & 119.7 & 825.1 & 403.1 \\ 
        ~ & $\epsilon$0.1 & 596.3 & 119.7 & 885.8 & 460.0 \\ 
        ~ & $\epsilon$=0.15 & 637.0 & 105.5 & 827.6 & 511.7 \\ 
        TS & ~ & 532.1 & 163.8 & 729.2 & 328.4 \\ 
        UCBPI & ~ & 534.9 & 100.6 & 771.3 & 415.4 \\ 
        UCB-Tuned & ~ & 707.7 & 175.8 & 1080.2 & 336.5 \\ 
        UCB & c=1 & 611.1 & 179.4 & 779.3 & 344.7 \\ 
        ~ & c=2 & 564.3 & 171.6 & 853.0 & 357.6 \\ \hline
    \end{tabular}
\end{table}

\begin{figure*}[h]
  \centering
  \includegraphics[width=1\linewidth,page=3]{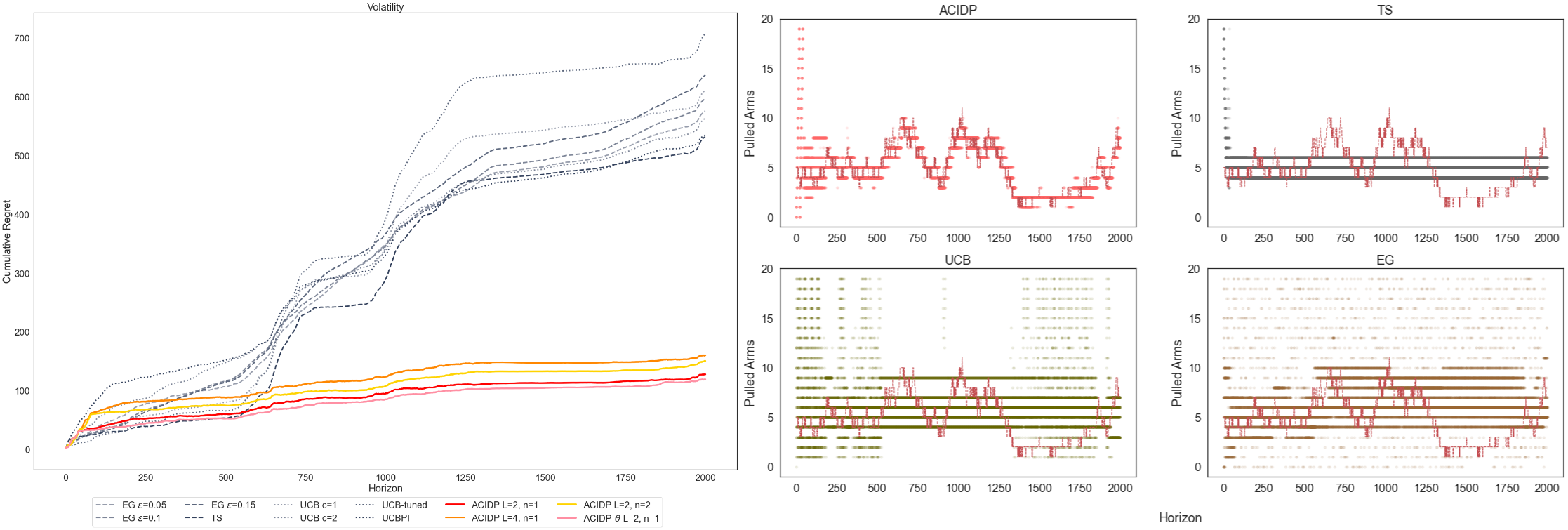}
  \caption{Regret and Price Selection (red line denotes $a^*$) in Case5: Volatility.}\label{fig:case5}
\end{figure*}

\clearpage

\subsection{Case6: Upside-Down}
The Upside-Down case is intended to imitate dramatic environmental shifts, and optimality can only be attained through a thorough exploration of the new environment. We designed it to be similar to the second case, which is more friendly to alternative policies, particularly EG. In this situation, while ACIDP recognizes upward pricing adjustments without transferability, they suffer from credibility degradation and do not completely transfer into the new environment. The regular ACIDP processes through all three steps of audition to achieve credibility, optimality, and transferability.

\begin{table}[h]
\caption{Case6 Result.}
    \centering
    \begin{tabular}{ll|llll}
    \hline
        Policy & Hyperparameters & Mean Regret & Standard Error & Max & Min \\ \hline
        ACIDP & L=2, n=1 & 509.0 & 503.0 & 1886.9 & 165.7 \\
        ~ & L=4, n=1 & 554.2 & 511.9 & 1926.0 & 217.7 \\ 
        ~ & L=2, n=2 & 365.7 & 126.1 & 629.7 & 221.5 \\ 
        ACIDP-$\theta$ & L=2, n=1 & 387.6 & 135.2 & 619.7 & 240.3 \\ 
        ACIDP & without transferability test & 1041.4 & 403.5 & 1911.0 & 597.0 \\ 
        ACIDP-w & window variant & 529.8 & 95.2 & 663.9 & 363.5 \\ 
        EG & $\epsilon$=0.05 & 698.9 & 241.8 & 1000.6 & 309.8 \\ 
        ~ & $\epsilon$=0.1 & 693.5 & 215.0 & 1074.8 & 379.2 \\ 
        ~ & $\epsilon$=0.15 & 933.5 & 218.9 & 1275.2 & 482.1 \\ 
        TS & ~ & 1197.7 & 198.6 & 1423.4 & 964.0 \\ 
        UCBPI & ~ & 1318.8 & 73.8 & 1519.8 & 1263.1 \\ 
        UCB-Tuned & ~ & 1453.8 & 208.6 & 1864.5 & 1113.9 \\ 
        UCB & c=1 & 1489.3 & 307.1 & 1931.1 & 895.6 \\ 
        ~ & c=2 & 1308.4 & 446.0 & 1679.4 & 292.6 \\ \hline
    \end{tabular}
    \label{table:ta2}
\end{table}

\begin{figure*}[h]
  \centering
  \includegraphics[width=1\linewidth,page=3]{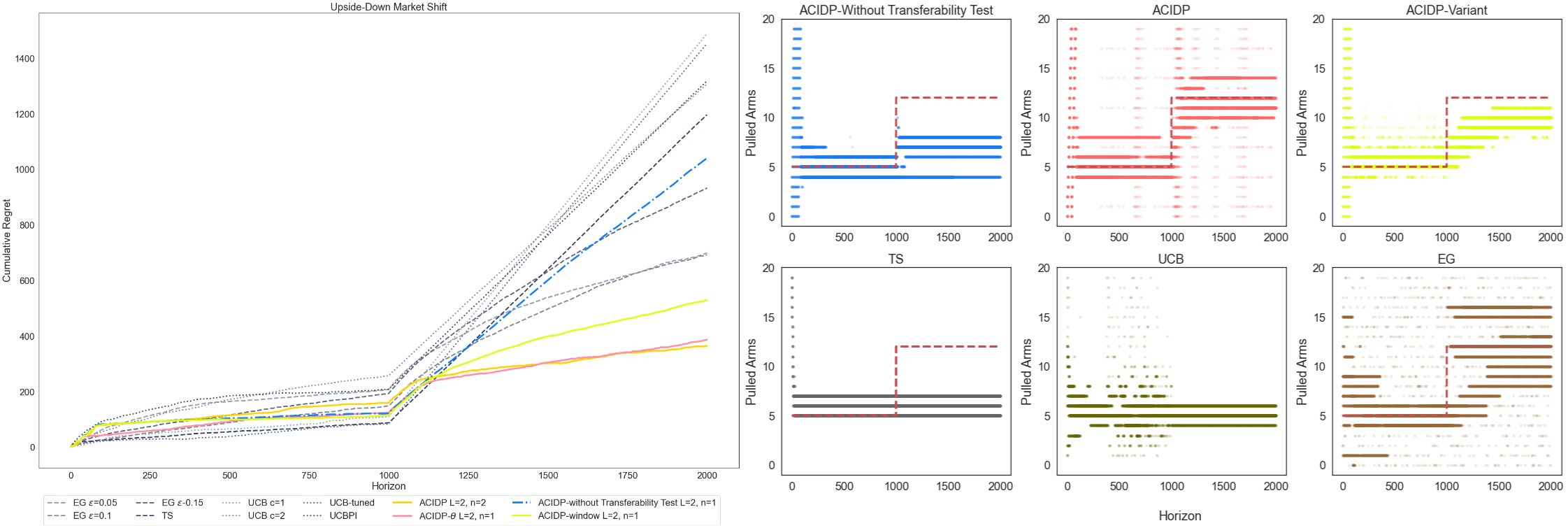}
  \caption{Regret and Price Selection (red line denotes $a^*$) in Case6: Upside-Down.}\label{fig:case6}
\end{figure*}

\clearpage

\subsection{Real World Data}
The Criteo Sponsored Search Conversion Log simulation is a real-life example of the Upside-Down case. This time the $T$, $k$, and $N$ are scaled up to cater to wider applications. Here we discussed several challenges identified in this dataset. Firstly, each of the three products has a completely distinct valuation and demand curve. The price needs to go up and down while the underlying demand are completely unrelated. Second, the valuation distribution cannot easily be approximated by common parametric families. This, however, rendered ACIDP an even more powerful policy as it does not assume any shape on valuation. Finally, conversion rates are typically low in the online advertisement industry (ranging from 2\% to 20\% across products.) Therefore, there are two critical factors affects N, number of clicks per round in this context. 1. how fast an agent is allowed to make a new decision. 2. the sample size needed in terms of demand utilized to update the decision. Because the typical number of clicks each day for a single product is roughly 250, we set N=500.\

\begin{table}[h]
\caption{Real Data: Criteo Experiment Result.}
    \centering
    \begin{tabular}{ll|llll}
    \hline
        Policy & Hyperparameters & Mean Regret & Standard Error & Max & Min \\ \hline
        ACIDP & L=4, n=1 & 38915.97 & 15637.81 & 78699.4 & 26595.33 \\ 
        ACIDP-w & window variant & 26723.16 & 2115.70 & 29568.53 & 23473.74 \\ 
        EG & $\epsilon$=0.05 & 83775.79 & 25410.65 & 127203.09 & 52124.14 \\ 
        ~ & $\epsilon$=0.1 & 106739.31 & 11245.25 & 120781.82 & 86027.40 \\ 
        ~ & $\epsilon$=0.15 & 133023.59 & 14774.32 & 158582.07 & 107559.25 \\ 
        TS & ~ & 199509.26 & 74.01 & 199622.38 & 199384.65 \\ 
        UCB-Tuned & ~ & 235209.70 & 88035.52 & 373923.95 & 132805.21 \\ 
        UCB & c=1 & 277109.95 & 139415.34 & 434026.10 & 107524.25 \\ 
        ~ & c=2 & 191249.57 & 101755.70 & 429739.28 & 107482.55 \\ \hline
    \end{tabular}
    \label{table:real}
\end{table}

\begin{figure*}[h]
  \centering
  \includegraphics[width=1\linewidth,page=3]{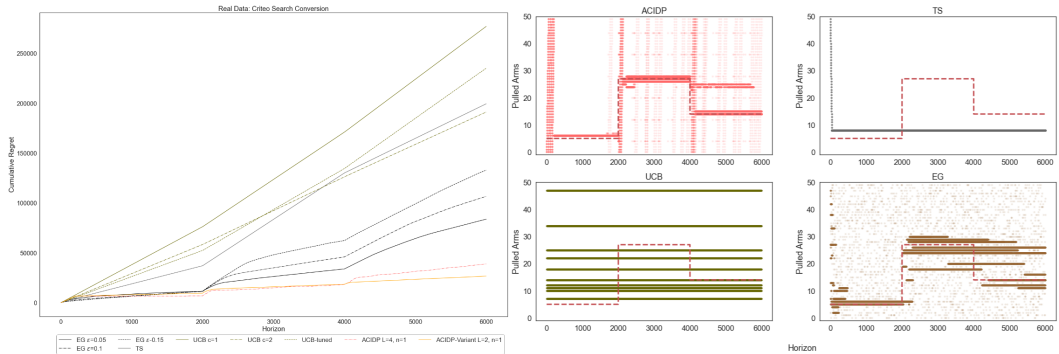}
  \caption{Regret and Price Selection (red line denotes $a^*$) in Real World Data.}\label{fig:criteo_2}
\end{figure*}

\clearpage

\section{Visualization of Experiment Details}
\textbf{Implementation of Algorithm3} Here we visualize the generator process in case6. The blue generated distribution is associated with $\theta^C$. It helps enhance credibility by approaching the true distribution. And in the estimated demand and profit curve, we show that ACIDP is capable of optimize price choice with limited and incomplete observation in both stages.

\begin{figure}[t]
  \centering
  \includegraphics[width=0.8\linewidth]{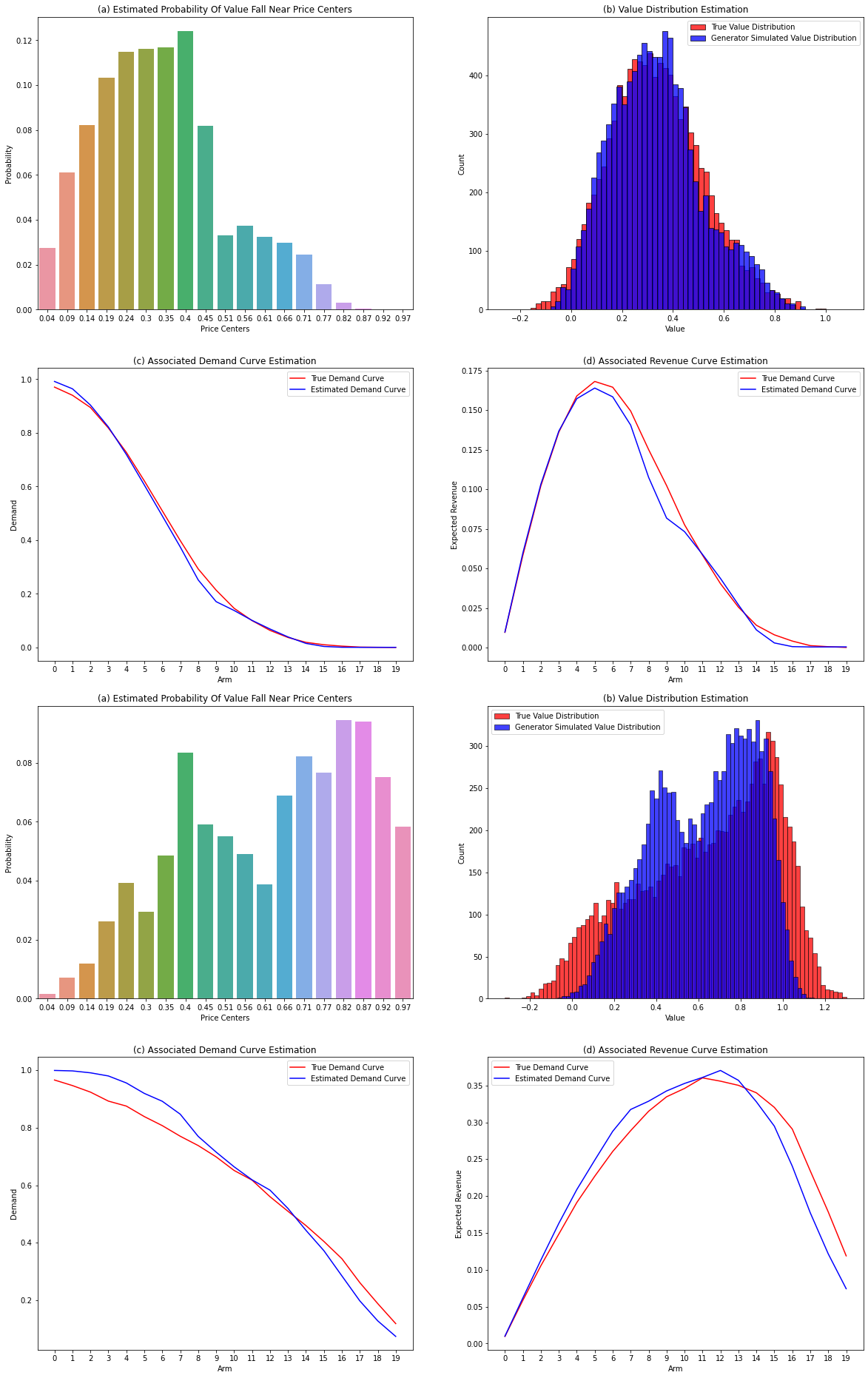}
  \caption{The counterfactual universe constructed before(above) and after(below) market shift in case 6. (a) discrete valuation derived from demand (b) decorated with noise (c) associated demand curve (d) associated profit curve}\label{fig:vis_gen1}
\end{figure}


\clearpage

\begin{figure}
  \centering
  \includegraphics[width=1\linewidth,page=3]{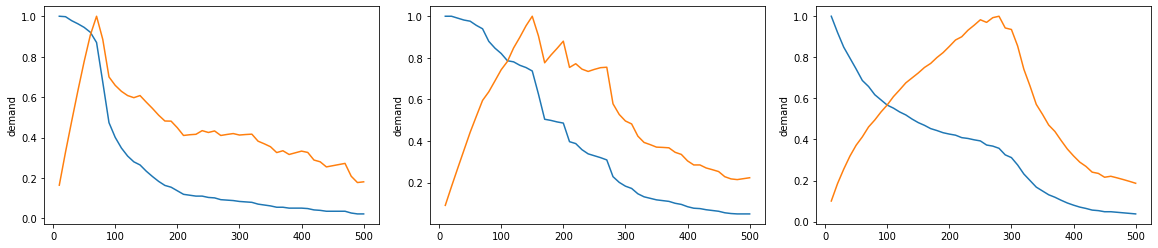}
  \caption{Normalized Demand (Blue) and Revenue (Orange) Curve of Product A, B, C.}\label{fig:productA}
\end{figure}

\textbf{Demand curve of the three products in Criteo Data}
The three chosen products A, B, C are originally indexed B442064C9D31A52BCAE0D18212FCFA1E, 0CAD35A37452D99F0BE65C9299EBF729, and 917676DCA7640B0256200468FB76B7BD in the dataset. Their demand curve is summarized as below. Their visualization in line plots (see \ref{fig:productA}) showcase how destructive real-life non-stationarity could be.\
\begin{table}
\caption{True Demand $D_{a_t}$ Calculated From The Criteo Dataset}
\begin{tabular}{l|lllll|lll} \hline
Price & Product A & Product B & Product C &  & Price & Product A & Product B & Product C \\ \hline
10    & 1         & 1         & 1         &  & 260   & 0.101     & 0.32      & 0.373     \\
20    & 0.998     & 1         & 0.921     &  & 270   & 0.093     & 0.309     & 0.368     \\
30    & 0.978     & 0.991     & 0.85      &  & 280   & 0.09      & 0.229     & 0.357     \\
40    & 0.963     & 0.982     & 0.796     &  & 290   & 0.088     & 0.201     & 0.325     \\
50    & 0.945     & 0.976     & 0.743     &  & 300   & 0.084     & 0.183     & 0.312     \\
60    & 0.921     & 0.956     & 0.687     &  & 310   & 0.081     & 0.172     & 0.276     \\
70    & 0.87      & 0.94      & 0.658     &  & 320   & 0.079     & 0.146     & 0.232     \\
80    & 0.674     & 0.879     & 0.618     &  & 330   & 0.07      & 0.132     & 0.2       \\
90    & 0.474     & 0.846     & 0.593     &  & 340   & 0.066     & 0.124     & 0.168     \\
100   & 0.401     & 0.821     & 0.567     &  & 350   & 0.062     & 0.117     & 0.149     \\
110   & 0.348     & 0.786     & 0.553     &  & 360   & 0.055     & 0.113     & 0.13      \\
120   & 0.308     & 0.781     & 0.534     &  & 370   & 0.055     & 0.11      & 0.119     \\
130   & 0.28      & 0.764     & 0.519     &  & 380   & 0.051     & 0.101     & 0.104     \\
140   & 0.264     & 0.753     & 0.499     &  & 390   & 0.051     & 0.095     & 0.091     \\
150   & 0.233     & 0.737     & 0.482     &  & 400   & 0.051     & 0.084     & 0.08      \\
160   & 0.207     & 0.625     & 0.469     &  & 410   & 0.048     & 0.077     & 0.071     \\
170   & 0.183     & 0.505     & 0.453     &  & 420   & 0.042     & 0.075     & 0.064     \\
180   & 0.163     & 0.499     & 0.443     &  & 430   & 0.04      & 0.069     & 0.056     \\
190   & 0.154     & 0.492     & 0.433     &  & 440   & 0.035     & 0.066     & 0.053     \\
200   & 0.137     & 0.486     & 0.426     &  & 450   & 0.035     & 0.062     & 0.048     \\
210   & 0.119     & 0.397     & 0.421     &  & 460   & 0.035     & 0.055     & 0.048     \\
220   & 0.115     & 0.388     & 0.409     &  & 470   & 0.035     & 0.051     & 0.045     \\
230   & 0.11      & 0.358     & 0.405     &  & 480   & 0.026     & 0.049     & 0.043     \\
240   & 0.11      & 0.338     & 0.398     &  & 490   & 0.022     & 0.049     & 0.04      \\
250   & 0.104     & 0.329     & 0.393     &  & 500   & 0.022     & 0.049     & 0.037    \\\hline
\end{tabular}
\end{table}

\end{appendices}
\end{document}